\documentclass[11pt,a4paper]{article}
\usepackage[hyperref]{emnlp2018}
\usepackage{times}
\usepackage{latexsym}

\usepackage{graphicx}
\usepackage{url}
\usepackage{xcolor}

\usepackage{natbib}

\usepackage{listings}
\usepackage{textcomp}

\aclfinalcopy % Uncomment this line for the final submission
\title{Patient Risk Assessment and Warning Symptom Detection Using Deep Attention-Based Neural Networks}

 \author{\\ \bf Ivan Girardi$^1$, Pengfei Ji$^{1,2}$, An-phi Nguyen$^1$, Nora Hollenstein$^{1,2}$, Adam Ivankay$^1$,\\ \bf Lorenz Kuhn$^1$, Chiara Marchiori$^1$ and Ce Zhang$^2$\\
             \normalsize $^1$ IBM Research Zurich, Switzerland\\
              \normalsize $^2$ ETH Zurich, Switzerland\\
\normalsize {\tt ivg@zurich.ibm.com, pji@student.ethz.ch, uye@zurich.ibm.com,}\\
\normalsize {\tt noraho@ethz.ch, aiv@zurich.ibm.com, kuhnl@student.ethz.ch,}\\ 
\normalsize {\tt chi@zurich.ibm.com, ce.zhang@ethz.ch} 
}

\date{}

\begin{document}
\maketitle
\begin{abstract}
We present an operational component of a real-world patient triage system. Given a specific patient presentation, the system is able to assess the level of medical urgency and issue the most appropriate recommendation in terms of best \textit{point of care} and \textit{time to treat}. We use an attention-based convolutional neural network
architecture trained on 600,000 doctor notes in German. We compare two approaches, one that uses the full text of the medical notes and one that uses only a selected list of 
medical entities extracted from the text. 
These approaches achieve 79\% and 66\% precision, respectively,
but on a confidence threshold of 0.6, precision increases
to 85\% and 75\%, respectively.
In addition, a method to detect \textit{warning symptoms} is implemented to render the classification task transparent from a medical perspective. The method is based on the learning of attention scores and a method of automatic validation using the same data. 
\end{abstract}

\section{Introduction}
Several intelligent triage systems have recently been developed that attempt to evaluate automatically the risk related to specific patient conditions and direct patients to the appropriate care provider \cite{semigran2015evaluation}. The work presented here is part of an interactive triage system being developed for industrial applications. 
The system takes patient demographics and symptoms as input, assesses their current medical conditions and suggests where and by when the patients should seek medical care. 
A key feature of the system is the detection of warning symptoms, namely, red flags. This is crucial to distinguish potential emergencies from common or less urgent cases and therefore provides the medical rationale behind a given recommendation. In addition, for triage systems that involve a dialogue with patients through multiple question-and-answer interactions (such as \citet{Ada}), warning symptom detection is fundamental to determine the most informative questions to ask patients.

We propose a model that assesses patient risk and detects warning symptoms based on a large volume of doctor notes in German, sometimes even mixed with Swiss German expressions. In this context, assessing patient risk can be regarded as a supervised text classification task, where the content of the medical records represents the feature space, and the recommendations assigned by medical professionals are the ground truth labels. The use of recurrent neural networks (RNN) has been proposed to solve text classification tasks \cite{tang2015document}.
However, the proposed RNN models must be modified
to be consistent with 
the requirement that warning symptoms must be detected, because in RNNs it is generally not possible to know which hidden states are most relevant. 

To address these challenges, we propose an integrated approach to assess patient risk and detect warning symptoms simultaneously using an attention-based convolutional neural network (ACNN), which is a combination of a convolutional neural network (CNN) and an attention mechanism \cite{kim2014convolutional,yang2016hierarchical,du2017convolutional}. To the best of our knowledge, such an integrated approach is applied for the first time to the medical domain.

The main contributions of this paper are twofold. First, we propose a neural network architecture that can be used simultaneously for text classification and the detection of important words. Comparing our model to other neural architectures of similar complexity, we achieve competitive classification results. The model is especially useful to explain the recommendation rationale in classification scenarios, where the given input consists of a set of extracted entities, rather than full text.
Second, a formal pipeline to detect warning symptoms based on learned importance factors is applied in an industrial application. Our model identifies symptoms that indicate a medical emergency. These warning symptoms can then be used by intelligent medical care services or in an ontology.

\section{Related Work}

\subsection{Text Classification with Deep Learning}

Traditional text classification approaches represent documents with sparse lexical features, such as $n$-grams, and use a linear model or kernel methods on this representation \cite{wang2012baselines,joachims1998text}.
More recently, deep learning technologies have been applied to text categorization problems.
RNNs are designed to handle sequences of any length and capture long-term dependencies. Like sequence-based \cite{tang2015document} and tree-structured \cite{tai2015improved} models, they have achieved remarkable results in document modeling.

Moreover, CNN models have achieved high accuracy on text categorization.
For example, \citet{kim2014convolutional} used one convolutional layer (with multiple widths and filters) followed by a max pooling layer over time.
\citet{johnson2015semi} built a model that uses up to six convolutional layers, followed by three fully connected classification layers.
\citet{conneau2016very} published a model with a 32-layer character-level CNN, that achieved a significant improvement on a large dataset. Models that combine CNN and RNN components for document classification also yield competitive results on several public datasets \cite{zhou2015c,lai2015recurrent}.

To the best of our knowledge, not many research efforts have focused on augmenting CNNs for text classification with attention mechanisms. 
In fact, attention layers are more typically coupled with RNNs in order to better handle long-term dependencies \cite{yang2016hierarchical}. Interestingly, \citet{du2017convolutional} used a CNN not as a classifier, but to compute the attention weights to apply to the hidden layers of a RNN. An example of combining attention layers with a CNN is the work by \citet{shen16attconv}. However, the authors do not augment the CNN features using attention weights. They use an attention mechanism to compute sentence-level features, which they then concatenate to the convolutional features to ultimately perform the classification.

\subsection{Intelligent Triage Systems}

Intelligent triage systems inform patients where and when they should seek medical care, based on methods such as expert rules, Bayesian inference and deep learning \cite{semigran2015evaluation}. For example, \citet{Symptomate} uses a Bayesian network and a medical database for triage advice. Clinical records written by medical experts have also been used to make triage suggestions with deep learning technologies. \citet{li2017convolutional} uses a shallow CNN model to predict a patient's diseases using the corresponding admission notes. \citet{nigam2016applying} applied a LSTM model to the multi-label classification task of assigning ICD-9 labels to medical notes.

\section{Methodology}

\subsection{Data Processing}

To build the triage application described here, we used 600,000
case records written in German and collected over the past five years. This is only 50\% of the total available data, as we selected 
only those cases treated by top-ranked doctors. 
Case records contain demographic data such as age and gender, previous illnesses, and a full-text description of the patient's current medical condition. Potential diagnoses consistent with the symptom description are listed.

The descriptions in the records are expressed in formal medical language as well as in layman's terminology. The notes are not always written in complete sentences and include misspellings, dialect vocabulary, non-standard medical abbreviations and inconsistent punctuation. This is a challenge for the linguistic processing of case files.

The original case records are very unevenly distributed over ten recommendation classes (a combination of a point-of-care
and a time-to-treat class). To mitigate this problem and for the purpose of this work, the original classes, (\textit{emergency, urgent}), (\textit{grundversorger, urgent}), (\textit{specialist, urgent}), (\textit{grundversorger, within a day}), (\textit{specialist, within a day}), (\textit{grundversorger, not urgent}), (\textit{specialist, not urgent}), (\textit{telecare, --}), were merged, with the help of healthcare professionals, into three categories: {\it Urgent Care, General Practice, Telecare}. The categorization of cases is shown in Table \ref{table:datap}.

\begin{table}[h!]
\centering
\begin{tabular}{|c c|} 
 \hline
 Recommendations& Number of Cases\\ [0.4ex] 
 \hline
 \hline
 Urgent Care & 270,000 \\
 General Practice & 104,000 \\
 Telecare & 244,000 \\

 \hline
\end{tabular}
\caption{Ground truth distribution for the reduced classes. \textit{Urgent Care} = Patient needs to seek medical care within a short time period; \textit{General Practice} = Patient requires medical attention in a physical consultation, but not urgently; \textit{Telecare} = In-person medical appointment not required, instructions over the phone are sufficient.}
\label{table:datap}
\end{table}

\subsubsection{NLP Pipeline}
\label{nlp}

A natural language processing (NLP) pipeline extracted medically relevant concepts associated with each written case. The pipeline consisted of the following stages: (1) data preprocessing for misspelling correction and
abbreviation expansion, (2) named entity recognition (NER) and (3) concept clustering. 
Acronyms and abbreviations used unambiguously were linked to the corresponding entities directly in the dictionaries. Ambiguous acronyms and abbreviations were resolved,
when possible, using algorithms that include context for disambiguation. 
For NER, we used a rule-based medical entity extraction system
built with IBM Watson Explorer, using algorithms based on dictionary look-up and advanced rules.
This allowed us to detect 51 entity types in the following categories: {\it anatomy, physiology, symptoms, diseases, 
medical procedures, medicines, negated symptoms, negated diseases, ability/inability of, foreign-body objects, negations, patient information, symptom characterization, disease
characterization, time expressions}.
The distinction between symptoms and diagnosis was made using existing ontologies, where these semantic types were assigned with the help of a team of clinical experts. The dictionaries used in the NER were built partially based on existing German-language medical dictionaries and ontologies (UMLS mapped German terms, ICD10, Meddra, etc.) and partially using the list of words contained in the case records. The dictionaries therefore contain a mapping of technical and layman's terms.
The NLP pipeline was designed to detect and resolve the negated mentions of the entities listed above (using
German language-specific negation particles or expressions), which are very frequent in this type of records. 
Only 31 entity types in the categories \textit{symptoms, diseases, ability/inability of,
negated symptoms, negated diseases}
were included in the current final list. 
The average number of extracted annotations per case was
70 for all entities, but only 17 for the selected entities. 
Performance was evaluated using the manual annotations of a set of ground truth cases performed by a team of clinical experts.
Concept clustering is a hierarchical procedure that allowed us to group annotations describing the same medical concept.
The same entity may be expressed in a variety of forms (compound vs. simple nouns, dialect or common language vs. medical terminology). Concept clustering is performed either at the dictionary level or by algorithms based on similarity between lemmas associated with the annotations.

Table~\ref{table:patient-case} lists the concepts extracted from an original case record after preprocessing by the described NLP pipeline.

\begin{table}[h!]
\centering
\begin{tabular}{l l}
\hline
\it key & \it value\\
\hline
\it Gender & \it Male\\[1.0ex]
\it Class & \it Urgent Care\\[1.0ex]
\it Content & \it ``Seit heute beschwert sich \\
& \it der Patient \"{u}ber heftige \\
& \it Brustschmerzen; Fieber 37,4\textdegree{}C;\\
& \it Schwierigkeiten beim Atmen,\\
& \it leichte Kopfschmerzen."\\[1.0ex]
\it Entities & \it starke Brustschmerzen\\
& \it Fieber 36-38\textdegree{}C\\
& \it Atembeschwerde\\
& \it leichte Kopfschmerzen\\[1.0ex]
\hline
\end{tabular}
\caption{(key,value)-pairs
of an original patient case file and
extracted entities.}
\label{table:patient-case}
\vspace{-0.2cm}
\end{table}

In this paper, we will benchmark the classification approach of using the extracted concepts with respect to the one of using the full text.

\subsection{Model Architecture}

\begin{figure*}[h!]
\centering
\vspace{-2.6cm}
\includegraphics[width=0.8\textwidth]{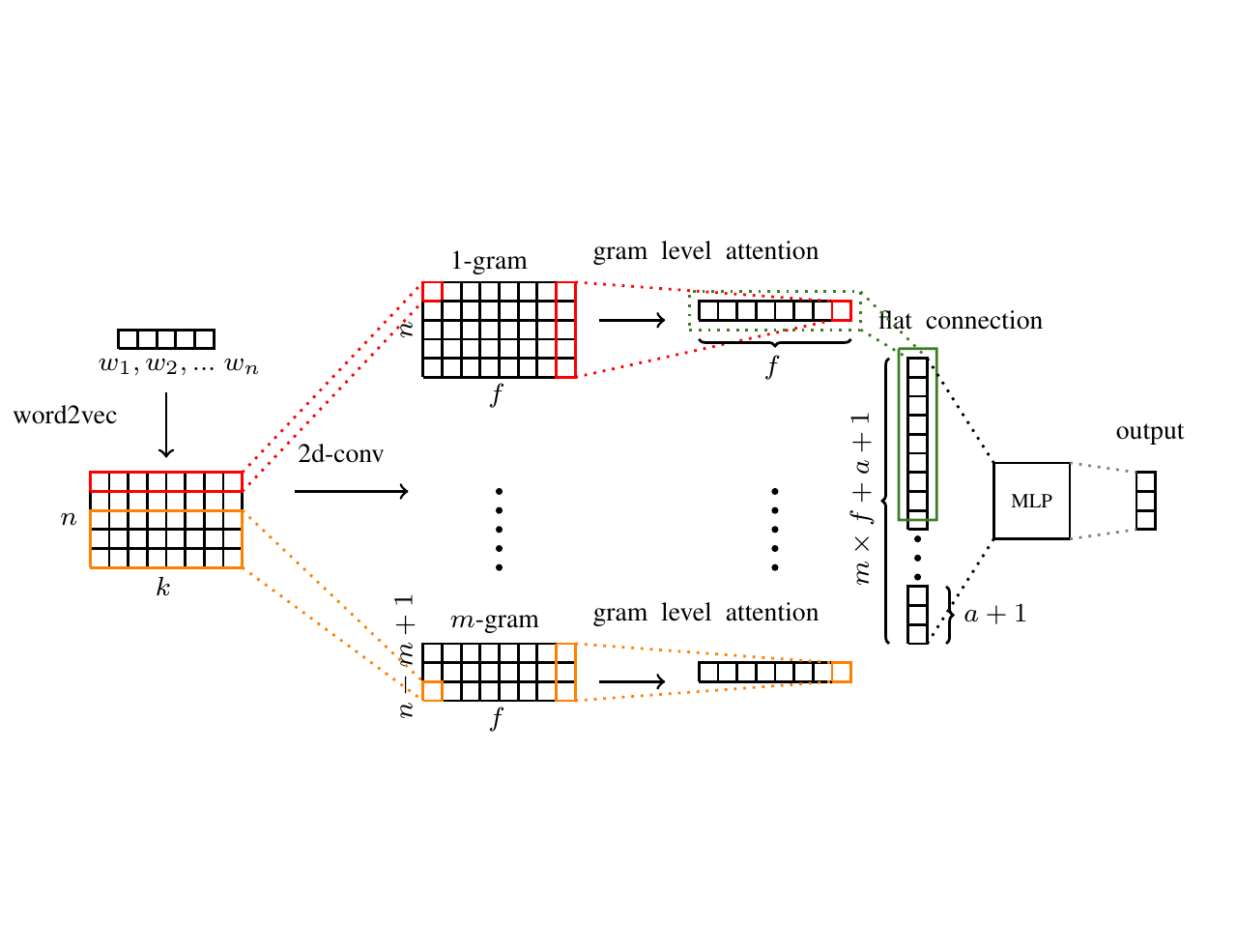}
\vspace{-2.5cm}
\caption{Model Architecture.}
\label{fig:architecture}
\end{figure*}

The overall architecture of the attention-based CNN is shown in Fig.~\ref{fig:architecture}. It consists of several components: a word embedding look-up layer obtained using \textit{word2vec} \cite{mikolov2013distributed}, a CNN-based $n$-gram encoder, an $n$-gram level attention layer and several fully-connected layers. By means of word embeddings, each word is represented as a real-valued vector. The word embedding look-up layer is a word embedding table \(T\in\mathcal{R}^{n \times k}\), where $n$ is the total vocabulary size and $k$ is the embedding dimension. The parameters of the embedding table were fine-tuned during the training phase.

\subsubsection{N-Gram Encoder}

We used a 2D convolution layer 
\cite{kim2014convolutional} to encode the word sequence into $n$-gram representations, thus capturing contextual information. For a given document, a 2D convolution filter \(w\in\mathcal R^{m \times k} \) was applied to a window of $m$ words to produce a new feature. A feature \(c_{i}\) was generated from a window of words \(x_{i:i+m-1}\) by 
\begin{equation}
    c_{i} = Relu(w\cdot x_{i:i+m-1} + b).\newline
\end{equation}
This filter was applied to each possible window of words in the sentence \({x_{1:m},x_{2:m+1},.....x_{n-m+1:n}}\) to produce a feature map:
\begin{equation}
    c = [c_{1},c_{2},......,c_{n-m+1}],
\end{equation}

\noindent with \(c \in\mathcal R^{n-m+1}\).
By applying multiple filters (denoted \(f\)) on \(x_{i:i+m-1}\), we obtained a new representation of the document. By setting different values for $m$, we obtained different $n$-gram representations of the documents. This operation was useful in our application setting because these layers create local region embeddings by $n$-grams.
Moreover, this allowed us to compute the attention factors for a combination of several symptoms. This in turn enabled us to detect pairs and even triplets of symptoms that are harmless if they appear individually, yet become red flags when they appear together. For example, the individual symptoms \textit{pain in arm} and \textit{sudden nausea} are no cause for concern. However, if a patient experiences both, this might indicate an impending heart attack. 

\subsubsection{N-Gram Level Attention Layer}
For each $n$-gram representation, we wanted to derive a corresponding fully-connected representation for the document. As different $n$-grams are of different importance to the document, we introduced an attention mechanism to extract $n$-grams that are relevant to the meaning of the document and aggregated the representation of those informative $n$-grams to form a document vector. The relevant $n$-grams then became candidates for warning symptoms. More specifically, the attention mechanism was defined such that:

\begin{equation}
    u_{it} = \tanh(W_{w}v_{it}+b_{w}),
\end{equation}

\noindent where \(v_{it}\) refers to the $t$th row of $i$th-gram representation. That is, we first fed the $n$-gram annotations \(v_{it}\) through a one-layer neural network to obtain \(u_{it}\) as a hidden representation of \(v_{it}\). Then we measured the importance of the word as the similarity of \(u_{it}\) with a word-level context vector \(u_{w}\) and obtained a normalized importance weight \(\alpha_{it}\) through a softmax function:

\begin{equation}
    \alpha_{it} = \frac{exp(u_{it}^Tu_{w})}{\sum_t exp(u_{it}^Tu_{w})}.
\end{equation}
The context vector \(u_{w}\) can be regarded as a high-level representation of a fixed query ``what is the most informative word?'' used in memory networks \cite{sukhbaatar2015end,kumar2016ask}. Context vector \(u_{w}\) was randomly initialized and jointly learned during the training process.  Thereafter, we computed the document vector \(s_i\) as a weighted sum of the $n$-gram annotations based on the weights:

\begin{equation}
    s_{i} =\sum_t\alpha_{it}v_{it}.
\end{equation}
Finally, all $n$-gram document level representations were flattened into a one-dimensional vector (flat connection layer in Fig.~\ref{fig:architecture}) plus patient gender and age ($a+1$ in Fig.~\ref{fig:architecture}). This vector was then fed into a multilayer perceptron (MLP) for classification. 

\subsection{Warning Symptom Detection}
\label{wsd_method}
Warning symptoms, or red flags, indicate the need for urgent medical care. 
The ACNN model is able to distinguish the importance of each symptom in the final classification.
Thereafter, we calculated the attention score for each symptom as follows:
\begin{eqnarray}
score(s_j) &=& \frac{\sum_{c_i \in C} \Phi(c_i,s_j) f(c_i,s_j)}{occur(s_j)},\\
f(c_i, s_j) &=&  \frac{att(s_j)}{ \max\limits_{s_k \in c_i}att(s_k)},
\end{eqnarray}
\noindent where $\Phi(c_i,s_j)$ is equal to 1 if symptom $s_j$ is contained in case record $c_i$
and zero elsewhere; \(C\) is the set of urgent care cases in the data; \(occur(s_{i})\) is the total occurrences of symptom \(s_{i}\); \(att(s_k)\) are the attention weights returned by the ACNN. The attention weights gave us a measurement of the warning level of the symptoms. 

This procedure was applied for all
classes to detect the most important symptoms that drive the model's prediction. As expected for the other classes,
the model assigns high attention weights to non-warning symptoms.

\section{Results}

\begin{table*}[h!]
\centering
\begin{tabular}{|l c c c c c c c c c |}
\hline
 Model & P($f_1$) & R($f_1$) & F($f_1$) & P($f_2$) & R($f_2$) & F($f_2$) & P($f_3$) & R($f_3$) & F($f_3$) \\ [0.5ex] 
 \hline\hline
 KIM CNN & 82.3 & 80.5 & 81.9 & 69.2 & 65.4 & 68.4 & 82.2 & 86.1 & 83.0 \\
 CLSTM & 78.2 & 82.6 & 79.0 & 66.9 & 62.4 & 66.0 & 83.4 & 80.7 & 82.8 \\
 BiGRU Attention Net & 74.9 & 80.2 & 75.9 & 62.6 & 59.0 & 61.9 & 80.8 & 76.6 & 79.9 \\
 ACNN & 80.5 & 81.1 & 80.7 & 67.6 & 60.9 & 66.1 & 82.0 & 84.8 & 82.6  \\ [0.2ex]
\hline
\end{tabular}
\caption{Prediction results in \% for the different architectures on \textit{full text}, where
P($f_k$), R($f_k$), F($f_k$) are precision, recall and f-score divided by class, and where $f_1$, $f_2$, $f_3$ are urgent care, general practice and telecare, respectively. Similar values were obtained by conducting several experiments and averaging the results.}
\label{table:Acc-full}
\end{table*}

\begin{table*}[h!]
\centering
\begin{tabular}{|l c c c c c c c c c |}
\hline
 Model & P($s_1$) & R($s_1$) & F($s_1$) & P($s_2$) & R($s_2$) & F($s_2$) & P($s_3$) & R($s_3$) & F($s_3$) \\ [0.5ex] 
 \hline\hline
 KIM CNN & 70.5 & 73.6 & 71.1 & 55.2 & 40.6 & 51.5 & 66.5 & 70.6 & 67.3 \\
 CLSTM & 70.0 & 71.6 & 70.3 & 53.8 & 40.1 & 50.4 & 65.4 & 70.8 & 66.4 \\
 BiGRU Attention Net & 69.2 & 72.5 & 69.9 & 53.0 & 43.1 & 50.7 & 66.7 & 68.4 & 67.0 \\
 ACNN & 72.5 & 68.2 & 71.6 & 51.9 & 47.8 & 51.0 & 65.5 & 72.0 & 66.5 \\ [0.2ex]
\hline
\end{tabular}
\caption{Same as Table \ref{table:Acc-full} but on \textit{symptoms} dataset, where $s_1$, $s_2$, $s_3$ are urgent care, general practice and telecare cases, respectively.}
\label{table:Acc-syn}
\end{table*}

\subsection{Patient Risk Assessment Experiment}

\subsubsection{Training Details}\label{train_det}
We conducted a detailed evaluation of this model on both the original \textit{full-text} dataset and a dataset of a few selected medical entities (see Section~\ref{nlp} for details) denoted for simplicity as a \textit{symptoms} dataset. The machine learning
framework where all the neural network models have been implemented was based on TensorFlow and Keras.
The vocabulary size, average document size and maximum document length are 134,000, $62.9$ and $959$ words for the full-text dataset; and 20,000, 14.15 and $94$ for the \textit{symptoms} dataset.
We used 90\% of the data for training, 5\% for validation, and 5\% for test randomly sampled.
Both datasets were preprocessed by removing stop words and low-occurrence words and zero-padding the documents. We learned 200-dimensional word embeddings on our datasets with \textit{word2vec} over 25 iterations. The embeddings were
different for each dataset.

We tuned our parameters on a 30,000 validation set and report the result on another 30,000 test set.
For model-specific parameters, we used grid search to find the optimal values.
We used a cross-entropy loss function with 256-mini-batch updating and Adam optimizer for five epochs. The learning rate was between 0.001 and 0.003; regularization was performed by weight decay of 0.0001 and a dropout of 0.8 was applied to every MLP layer. The attention vector size was set up to 100, and the window size was set from 1 to 5. For each $n$-gram extraction, we used up to 128 filters for 2D convolution.

\subsubsection{Model Comparison}

In this section, we compare our system to the following approaches:

\noindent\begin{bf}CLSTM\end{bf} \cite{zhou2015c} applies a CNN model on text and feeds consecutive window features directly to a LSTM model.

\noindent\begin{bf}Kim CNN\end{bf} \cite{kim2014convolutional} uses 2D convolution windows to extract an $n$-gram representation followed by max-pooling. 

\noindent \begin{bf}BiGRU Attention Network\end{bf} \cite{yang2016hierarchical} consists of RNNs applied on both word and sentence level to extract a hidden state. An attention mechanism is applied after the bidirectional gated recurrent units.

The results on the datasets with the {\it full text} and the {\it symptoms} only are shown in Tables~\ref{table:Acc-full} and \ref{table:Acc-syn}, respectively.
All the analyzed models show similar performance in the classification task. For all models, the performance decreases as we move from the full text dataset to the symptoms dataset because the medical and contextual information also diminishes by taking into account only the extracted symptom concepts.  

\subsubsection{Result Analysis}

\begin{figure*}[h!]
\includegraphics[width=\textwidth]{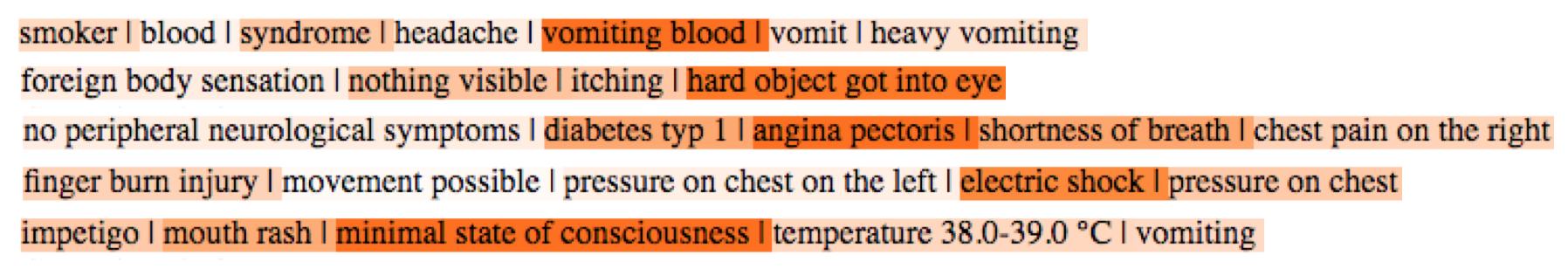}
\caption{Visualization of attention factors from neural network used to explain recommendation rationale. Each line represents the (translated) symptoms extracted for a patient case file. The darker the color, the higher the attention factor for a symptom.}\label{att_values}
\end{figure*}

In this section, we compare our ACNN model with the
state-of-the-art deep learning models to obtain a benchmark
on our triage use case.
We also describe how our approach, a combination of convolutional neural networks and attention mechanisms, equals the performance of existing models with the
advantage of being explainable.

Kim CNN uses 2D convolution windows to extract $n$-gram representations. Max pooling was then applied to each of the filter outputs. A single value was retained for each feature map. This might work well for short sentences containing only a few ``leading'' words indicating the category. For longer documents, however, all information about $n$-grams is lost apart from the strongest signal. The presence of highly important symptoms in clinical data is the reason why this model performs well especially for  urgent care and telecare classes. This hypothesis is supported by the number of symptoms with large attention scores found in the ACNN model for these classes.

The BiGRU Attention Network applies an attention layer after bidirectional GRU components. For a given word in a sentence, it encodes information about the word context in that sentence. However, compared to a 2D convolution window, only a single context window is used. It is not trivial to choose the optimal window size. Thus, it is difficult to detect warning symptom pairs or triplets. For 2D convolution in our model, identifying such pairs or triplets would be more straightforward because attention factors are also learned for $2$ and $3$-grams. Another limitation of GRU models is that they rely on fully sequential data. In our use case, however, the data is composed of several separate phrases, words or incomplete sentences.

Our ACNN combines the merits of 2D convolution and attention mechanisms by stacking 2D convolution layers to extract contextual information and an attention mechanism to assign importance factors to different symptoms and combinations thereof.

\subsection{Warning Symptom Detection}\label{eval_wsd}
Owing to the lack of ground truth, we used the following evaluation method to detect warning symptoms with the ACNN. First, we measured the recall of the ACNN on urgent care cases containing only symptom concepts. Then, a new dataset was created by removing from each case record the $1$-gram with the highest attention score, calculated as described in Section~\ref{wsd_method}. For urgent cases, we expected the removed $1$-grams to be highly important signals of medical urgency, hence warning symptoms. For instance, \textit{starke Brustschmerzen} would be removed from the case described in Section~\ref{nlp}. We then compared the ACNN recall for the urgent cases on the new dataset (Attention Drop) with respect to the recall on the original symptoms dataset (Baseline). This procedure is performed on all the classes
for validation.
The decrease in recall demonstrates the importance of the detected warning symptoms in order to classify urgent cases correctly. To verify that the detected warning symptoms are indeed highly informative, we furthermore generated datasets in which either random symptoms (Random Drop) or symptoms that appear most frequently in urgent cases (Frequency Drop) are dropped.

\begin{table*}[h!]
\centering
\begin{tabular}{|l c c c c c c c c c|} 
 \hline
 Dataset & P($s_1$) & R($s_1$) & F($s_1$) & P($s_2$) & R($s_2$) & F($s_2$) & P($s_3$) & R($s_3$) & F($s_3$)\\ [0.4ex] 
 \hline
 \hline
Baseline & 72.5 & 68.2 & 71.6 & 51.9 & 47.8 & 51.0 & 65.5 & 72.0 & 66.7 \\ 
Random Drop & 71.7 & 65.7 & 70.4 & 49.9 & 44.2 & 48.6 & 65.8 & 72.8 & 67.1\\
Frequency Drop & 71.7 & 65.7 & 70.4 & 50.9 & 46.0  & 49.8 & 66.0 & 73.5 & 67.4\\
Attention Drop & 70.3 & 61.3 & 68.2 & 48.0 & 40.9 & 46.4 & 66.3 & 74.6 & 67.8 \\
2 Random Drops & 70.8 & 62.8 & 69.0 & 47.6 & 40.2 & 45.9 & 66.0 & 73.3 & 67.3\\
2 Frequency Drops & 70.3 & 61.2 & 68.3 & 49.0 & 42.6 & 47.6 & 66.6 & 75.6 & 68.2 \\
2 Attention Drops & 67.5 & 53.6 & 64.2 & 44.0 & 34.8 & 41.8 & 67.0 & 77.0 & 68.8\\
 \hline
\end{tabular}
\caption{Different datasets and the model's precision, recall and f-score in \%, where $s_1$, $s_2$, $s_3$ are urgent care, general practice and telecare {\it symptoms} dataset, respectively.}
\label{wsd_res3}
\end{table*}
As shown in Table~\ref{wsd_res3}, dropping the attention-detected warning symptoms led to the largest decrease in performance. The difference became even more distinct if two symptoms instead of one were removed from the cases. 

Performance also decreased for the urgent care and general practice classes, whereas almost a flat
behavior was found for telecare class, as expected. In the latter case,
random, frequency, attention drops showed the same results because several features had the same attention scores.
Manual inspection of the symptoms with the highest attention scores
further supports these results.
The darker the color of the symptom in Figure \ref{att_values}, the higher its attention factor in the model. In the examined samples, darker colors did indeed correlate with symptoms that made patients require urgent care, such as \textit{vomiting blood} and \textit{electric shock}.

With single or double removals for the {\it full-text} dataset, a much lower decrease in performance was observed because of the higher number of features per case.

\subsection{Explainable Deep Learning}\label{expl_dl}

In current research, but especially in medical industry applications, \textit{transparent} or \textit{explainable} machine learning models are becoming increasingly important. 
Some machine learning models
have become so complex, they are black boxes.
End users need to understand why  a  certain  recommendation  was made.

In our application, the attention mechanism on which we based our warning (and non-warning) symptom detection represents a transparent method of reasoning why a given case belongs to a certain class. 

For instance, by analyzing the patient symptoms with the highest attention scores, it becomes apparent \textit{why} a case would be predicted to be urgent, general practice or telecare. Table \ref{high-scores} shows some examples with high/low attention scores computed using 1-gram attention values for urgent care, general practice and telecare classes. As can be seen, the symptoms with the highest score in the urgent cases are the most severe, whereas the symptoms in the telecare cases are less severe. In other words, symptoms with a high/low score for a given class are the most/least relevant ones for that class.
As expected, if the model predicts an urgent (non-urgent) class,
the model assigns a higher weight to warning (non-warning) symptoms.
The computation of 1-gram feature scores results in 2,000 (3,600), 734 (3,700), 1,500 (3,800) features with scores of $> 0.8$ ($< 0.2$) for $s_1$, $s_2$ and $s_3$, respectively.
The use of an attention layer on $n$-gram representations
allowed us to compute feature relevance including correlations 
between pairs, triplets, etc. An example of scores of feature pairs 
obtained by extracting the attention weights for the 2-grams is 
shown in Tables~\ref{high-scores-bigrams-1} and \ref{high-scores-bigrams-2}. Strong correlation
between feature pairs is found for the cases where
the score of the pair is much higher than those
of the single features.
The computation of 2-gram feature scores results in 12,000 (28,000), 4,800 (13,000), 10,000 (24,000) features with scores of $> 0.8$ ($< 0.2$) for $s_1$, $s_2$ and $s_3$, respectively.

\begin{table*}[h!]
\centering
\small
\begin{tabular}{|l l | l l |  l l |} 
 \hline
$s_1$ & score & $s_2$ & score & $s_3$ & score \\ [0.4ex] 
 \hline
 \hline

shortness of breath & 1.0 & intermittent shoulder pain & 1.0 & back distortion & 1.0\\ 
\multicolumn{2}{|l|}{(\textit{\small{Atemnot}})} &   \multicolumn{2}{l|}{(\textit{\small{intermittierende Schulterschmerzen}})} & \multicolumn{2}{l|}{(\textit{\small{R\"{u}ckenzerrung}})}\\[0.5ex]
pain after accident & 1.0 & severely itchy wound &	1.0 & abrasion on the back &	1.0 \\
\multicolumn{2}{|l|}{(\textit{\small{Schmerzen nach Unfall}})} &   \multicolumn{2}{l|}{(\textit{\small{stark juckende Wunde}})} & \multicolumn{2}{l|}{(\textit{\small{Sch\"{u}rfung am R\"{u}cken}})}\\[0.5ex]
foreign body in esophagus & 1.0 & purulent nasal discharge &	1.0 & toenail injury &	1.0 \\
\multicolumn{2}{|l|}{(\textit{\small{Fremdk\"{o}rper im \"{O}sophagus}})} &   \multicolumn{2}{l|}{(\textit{\small{eitriger Nasenausfluss}})} & \multicolumn{2}{l|}{(\textit{\small{Zehennagelverletzung}})}\\[0.5ex]
severe rectal bleed & 1.0 & neck abscess &	1.0 & itching forehead &	1.0 \\
\multicolumn{2}{|l|}{(\textit{\small{blutet stark rektal}})} &   \multicolumn{2}{l|}{(\textit{\small{Abszess am Nacken}})} & \multicolumn{2}{l|}{(\textit{\small{Juckreiz an der Stirn}})}\\[0.5ex]
itching back & 0.05 & no pain when walking &	0.03 & stabbed with knife &	0.03 \\
\multicolumn{2}{|l|}{(\textit{\small{Juckreiz am R\"{u}cken}})} &   \multicolumn{2}{l|}{(\textit{\small{beim Laufen keine Schmerzen}})} & \multicolumn{2}{l|}{(\textit{\small{Messerstich}})}\\[0.5ex]
pain in thumb & 0.04 & throat is normal &	0.01 & ear bleeding &	0.02 \\
\multicolumn{2}{|l|}{(\textit{\small{Daumenschmerz}})} &   \multicolumn{2}{l|}{(\textit{\small{Hals normal}})} & \multicolumn{2}{l|}{(\textit{\small{Ohrenblutung}})}\\[0.5ex]
nail injury & 0.03 & blister on tongue &	0.01 & hardened lower abdomen &	0.01 \\
\multicolumn{2}{|l|}{(\textit{\small{Nagelverletzung}})} &   \multicolumn{2}{l|}{(\textit{\small{Blase auf der Zunge}})} & \multicolumn{2}{l|}{(\textit{\small{verh\"{a}rteter Unterbauch}})}\\[0.5ex]
wart on foot & 0.01 & can drink normally &	0.003 & difficulty breathing &	0.005 \\
\multicolumn{2}{|l|}{(\textit{\small{Warze am Fuss}})} &   \multicolumn{2}{l|}{(\textit{\small{kann normal trinken}})} & \multicolumn{2}{l|}{(\textit{\small{Schwierigkeiten beim Atmen}})}\\
\hline
\end{tabular}
\caption{Symptoms (translated from German into English) scores divided by class using 1-gram attention values (only a few examples with high/low scores shown here). The corresponding German terms are given in parentheses.
}
\label{high-scores}
\end{table*}

\begin{table*}[h!]
\centering
\begin{tabular}{|l l l l |} 
 \hline
$(f_i$, $f_j)$ & score of $f_i$ & score of $f_j$ & score of $(f_i, f_j)$ \\ [0.4ex] 
 \hline
 \hline
(acute abdominal pain, severe abdominal pain) & 0.86 & 0.39 & 1.0\\
\multicolumn{4}{|l|}{(\textit{akute Bauchschmerzen, starke Bauchschmerzen})}\\[0.5ex]
(loss of consciousness, head injury) & 0.32 & 0.55 & 1.0\\
\multicolumn{4}{|l|}{(\textit{Bewusstseinsverlust, Sch\"{a}delverletzungen})}\\[0.5ex]
(epigastric pain, colic) & 0.35 & 0.26 & 1.0\\
\multicolumn{4}{|l|}{(\textit{Oberbauchschmerzen, Kolik})}\\[0.5ex]
(pneumonia, respiratory tract inflammation) & 0.45 & 0.24 & 0.92\\
\multicolumn{4}{|l|}{(\textit{Pneumonie, Atemwegentz\"{u}ndung})}\\[0.5ex]
(severe vomiting, dehydration) & 0.44 & 0.56 & 0.87\\
\multicolumn{4}{|l|}{(\textit{starkes Erbrechen, Dehydration})}\\[0.5ex]
(very high blood pressure, hypertensive crisis) & 0.49 & 0.78 & 0.86\\
\multicolumn{4}{|l|}{(\textit{Blutdruck stark erh\"{o}ht, hypertensive Krise})}\\[0.2ex]
 \hline
\end{tabular}
\caption{Symptoms (translated from German into English) scores
using 2-gram attention values (only a few high scores shown here) for $s_1$. The corresponding German terms are given in parentheses.
}
\label{high-scores-bigrams-1}
\end{table*}

\begin{table*}[h!]
\centering
\begin{tabular}{|l l l l |} 
  \hline
$(f_i$, $f_j)$ & score of $f_i$ & score of $f_j$ & score of $(f_i, f_j)$ \\ [0.4ex] 
 \hline
 \hline
(chronic back pain, back pain) & 0.29 & 0.08 & 1.0\\
\multicolumn{4}{|l|}{(\textit{chronische R\"{u}ckenschmerzen, R\"{u}ckenschmerzen})}\\[0.5ex]
(patella pain, knee pain) & 0.26 & 0.21 & 0.79\\
\multicolumn{4}{|l|}{(\textit{Schmerzen an der Kniescheibe, Knieschmerzen})}\\[0.5ex]
(chronic anemia, food allergy) & 0.18 & 0.17 & 0.66\\
\multicolumn{4}{|l|}{(\textit{chronische An\"{a}mie, Nahrungsmittelallergie})}\\[0.5ex]
(rheumatoid arthritis, joint pain) & 0.13 & 0.13 & 0.59\\
\multicolumn{4}{|l|}{(\textit{rheumatoide Arthritis, Gelenkschmerzen})}\\[0.5ex]
(colonoscopy, blood in stool) & 0.22 & 0.22 & 0.59\\
\multicolumn{4}{|l|}{(\textit{Darmspiegelung, Blut im Stuhlgang})}\\[0.5ex]
(fatigue, chronic anemia) & 0.07 & 0.18 & 0.58\\
\multicolumn{4}{|l|}{(\textit{M\"{u}digkeit, chronische An\"{a}mie})}\\[0.2ex]
 \hline
 \hline
(non-swollen lymph nodes, viral infection) & 0.20 & 0.31 & 1.0\\
\multicolumn{4}{|l|}{(\textit{keine Lymphknotenschwellung, virale Entz\"{u}ndung})}\\[0.5ex]
(conjunctivitis, slight redness) & 0.32 & 0.17 & 1.0\\
\multicolumn{4}{|l|}{(\textit{Konjunktivitis, leichte R\"{o}tung})}\\[0.5ex]
(abnormally frequent urination, no complication) & 0.26 
& 0.63 & 1.0\\
\multicolumn{4}{|l|}{(\textit{häufiges Urinieren, keine Komplikation})}\\[0.5ex]
(bladder infection, no pregnancy) & 0.29 & 0.21 & 0.96\\
\multicolumn{4}{|l|}{(\textit{Harnblasenentz\"{u}ndung, keine Schwangerschaft})}\\[0.5ex]
(local reaction, itchiness) & 0.31 & 0.17 & 0.92\\
\multicolumn{4}{|l|}{(\textit{Lokalreaktion, Juckreiz})}\\[0.5ex]
(gastroenteritis, no travel abroad) & 0.42 & 0.23 & 0.90\\
\multicolumn{4}{|l|}{(\textit{Gastroenteritis, kein Auslandaufenthalt})}\\[0.2ex]
 \hline
\end{tabular}
\caption{Symptoms (translated from German into English) scores divided by class using 2-gram attention values (only a few high scores shown here) for $s_2$ (upper), $s_3$ (lower) panel. The corresponding German terms are given in parentheses. 
}
\label{high-scores-bigrams-2}
\vspace{-0.1cm}
\end{table*}

\subsection{Confidence}\label{conf_dl}
To reach higher performance in an operative triage application, we define a confidence score in the classification based on which
the system decides whether to trust the recommendation.
In Table~\ref{table:Acc-full-thr}
and Table~\ref{table:Acc-syn-thr} we show the same results obtained in
Tables~\ref{table:Acc-full} and \ref{table:Acc-syn}, respectively,
discarding all test cases in which the predicted probability 
of the classifier was lower than $0.6$. With the chosen threshold,
we discarded roughly 30\% cases. Overall a performance improvement of between 5\% and 10\% is observed. In future work, we plan to apply additional techniques, e.g., based on hierarchical decision trees, to minimize medical risk even further.

\begin{table*}[h!]
\centering
\begin{tabular}{|l c c c c c c c c c |}
\hline
 Model & P($f_1$) & R($f_1$) & F($f_1$) & P($f_2$) & R($f_2$) & F($f_2$) & P($f_3$) & R($f_3$) & F($f_3$) \\ [0.5ex] 
 \hline\hline
 KIM CNN & 87.8 & 86.3 & 87.5 & 73.0 & 78.6 & 74.0 & 90.6 & 90.0 & 90.4 \\
 CLSTM & 84.4 & 88.4 & 85.2 & 76.2 & 66.5 & 74.0 & 88.4 & 87.6 & 88.3 \\
 BiGRU Attention Net & 76.5 & 82.2 & 77.6 & 65.1 & 60.7 & 64.2 & 82.5 & 78.2 & 81.6 \\
 ACNN & 85.3 & 87.3 & 85.6 & 77.3 & 65.1 & 74.5 & 87.1 & 89.5 & 87.6  \\ [0.2ex]
\hline
\end{tabular}
\caption{Same as Table \ref{table:Acc-full} applying a threshold to the probabilities of $0.6$.}
\label{table:Acc-full-thr}
\vspace{-0.1cm}
\end{table*}

\begin{table*}[h!]
\centering
\begin{tabular}{|l c c c c c c c c c |}
\hline
 Model & P($s_1$) & R($s_1$) & F($s_1$) & P($s_2$) & R($s_2$) & F($s_2$) & P($s_3$) & R($s_3$) & F($s_3$) \\ [0.5ex] 
 \hline\hline
 KIM CNN & 75.6 & 86.0 & 77.5 & 65.0 & 45.5 & 59.8 & 77.5 & 72.0 & 76.3 \\
 CLSTM & 76.5 & 83.2 & 77.8 & 66.8 & 44.5 & 60.7 & 75.2 & 75.2 & 75.2 \\
 BiGRU Attention Net & 73.2 & 76.8 & 73.4 & 58.4 & 45.1 & 55.1 & 70.5 & 72.5 & 70.9 \\
 ACNN & 77.0 & 81.5 & 77.9 & 62.8 & 60.0 & 60.3 & 75.7 & 74.9 & 75.5 \\ [0.2ex]
\hline
\end{tabular}
\caption{Same as Table \ref{table:Acc-syn} applying a threshold to the probabilities of $0.6$.}

\label{table:Acc-syn-thr}
\vspace{-0.3cm}
\end{table*}

\section{Conclusion}

We have described an attention-based CNN model to assess patient risk and to detect warning symptoms, which will be used in an industrial application for medical triage. We achieved a precision of 
79\% on the \textit{full-text} dataset and 66\% on the \textit{symptoms} set.
On a confidence threshold of 0.6, precision increases 
to 85\% and 75\%, respectively.
The learned attention weights allowed us to compute the symptom relevance, i.e., the attention score, which is then used to extract warning symptoms more precisely and to make the recommendation rationale transparent.
\\
\\
{\bf Acknowledgements}. We deeply acknowledge D.~Dykeman, D.~Ortiz-Yepes and K.~Thandiackal.

\newpage

%Include your own bib file like this:
\bibliographystyle{acl_natbib_nourl}
\bibliography{emnlp2018}

\begin{thebibliography}{20}
\expandafter\ifx\csname natexlab\endcsname\relax\def\natexlab#1{#1}\fi

\bibitem[{Ada(2018)}]{Ada}
Ada. 2018.
\newblock https://ada.com.

\bibitem[{Conneau et~al.(2016)Conneau, Schwenk, Barrault, and
  Lecun}]{conneau2016very}
Alexis Conneau, Holger Schwenk, Lo{\"\i}c Barrault, and Yann Lecun. 2016.
\newblock Very deep convolutional networks for natural language processing.
\newblock \emph{arXiv preprint arXiv:1606.01781}.

\bibitem[{Du et~al.(2017)Du, Gui, Xu, and He}]{du2017convolutional}
Jiachen Du, Lin Gui, Ruifeng Xu, and Yulan He. 2017.
\newblock A convolutional attention model for text classification.
\newblock In \emph{National CCF Conference on Natural Language Processing and
  Chinese Computing}, pages 183--195. Springer.

\bibitem[{Joachims(1998)}]{joachims1998text}
Thorsten Joachims. 1998.
\newblock Text categorization with support vector machines: Learning with many
  relevant features.
\newblock \emph{Machine learning: ECML-98}, pages 137--142.

\bibitem[{Johnson and Zhang(2015)}]{johnson2015semi}
Rie Johnson and Tong Zhang. 2015.
\newblock Semi-supervised convolutional neural networks for text categorization
  via region embedding.
\newblock In \emph{Advances in neural information processing systems}, pages
  919--927.

\bibitem[{Kim(2014)}]{kim2014convolutional}
Yoon Kim. 2014.
\newblock Convolutional neural networks for sentence classification.
\newblock \emph{arXiv preprint arXiv:1408.5882}.

\bibitem[{Kumar et~al.(2016)Kumar, Irsoy, Ondruska, Iyyer, Bradbury, Gulrajani,
  Zhong, Paulus, and Socher}]{kumar2016ask}
Ankit Kumar, Ozan Irsoy, Peter Ondruska, Mohit Iyyer, James Bradbury, Ishaan
  Gulrajani, Victor Zhong, Romain Paulus, and Richard Socher. 2016.
\newblock Ask me anything: Dynamic memory networks for natural language
  processing.
\newblock In \emph{International Conference on Machine Learning}, pages
  1378--1387.

\bibitem[{Lai et~al.(2015)Lai, Xu, Liu, and Zhao}]{lai2015recurrent}
Siwei Lai, Liheng Xu, Kang Liu, and Jun Zhao. 2015.
\newblock Recurrent convolutional neural networks for text classification.
\newblock In \emph{AAAI}, volume 333, pages 2267--2273.

\bibitem[{Li et~al.(2017)Li, Konomis, Neubig, Xie, Cheng, and
  Xing}]{li2017convolutional}
Christy Li, Dimitris Konomis, Graham Neubig, Pengtao Xie, Carol Cheng, and Eric
  Xing. 2017.
\newblock Convolutional neural networks for medical diagnosis from admission
  notes.
\newblock \emph{arXiv preprint arXiv:1712.02768}.

\bibitem[{Mikolov et~al.(2013)Mikolov, Sutskever, Chen, Corrado, and
  Dean}]{mikolov2013distributed}
Tomas Mikolov, Ilya Sutskever, Kai Chen, Greg Corrado, and Jeffrey Dean. 2013.
\newblock Distributed representations of words and phrases and their
  compositionality.
\newblock In \emph{Advances in neural information processing systems}, pages
  3111--3119.

\bibitem[{Nigam(2016)}]{nigam2016applying}
Priyanka Nigam. 2016.
\newblock Applying deep learning to {ICD}-9 {M}ulti-label {C}lassification from
  {M}edical {R}ecords.
\newblock Technical report, Stanford University.

\bibitem[{Semigran et~al.(2015)Semigran, Linder, Gidengil, and
  Mehrotra}]{semigran2015evaluation}
Hannah~L Semigran, Jeffrey~A Linder, Courtney Gidengil, and Ateev Mehrotra.
  2015.
\newblock Evaluation of symptom checkers for self diagnosis and triage: audit
  study.
\newblock \emph{bmj}, 351:h3480.

\bibitem[{Shen and Huang(2016)}]{shen16attconv}
Yatian Shen and Xuanjing Huang. 2016.
\newblock Attention-based convolutional neural network for semantic relation
  extraction.
\newblock In \emph{{COLING} 2016, 26th International Conference on
  Computational Linguistics, Proceedings of the Conference: Technical Papers,
  December 11-16, 2016, Osaka, Japan}, pages 2526--2536.

\bibitem[{Sukhbaatar et~al.(2015)Sukhbaatar, Weston, Fergus
  et~al.}]{sukhbaatar2015end}
Sainbayar Sukhbaatar, Jason Weston, Rob Fergus, et~al. 2015.
\newblock End-to-end memory networks.
\newblock In \emph{Advances in neural information processing systems}, pages
  2440--2448.

\bibitem[{Symptomate(2018)}]{Symptomate}
Symptomate. 2018.
\newblock https://symptomate.com.

\bibitem[{Tai et~al.(2015)Tai, Socher, and Manning}]{tai2015improved}
Kai~Sheng Tai, Richard Socher, and Christopher~D Manning. 2015.
\newblock Improved semantic representations from tree-structured long
  short-term memory networks.
\newblock \emph{arXiv preprint arXiv:1503.00075}.

\bibitem[{Tang et~al.(2015)Tang, Qin, and Liu}]{tang2015document}
Duyu Tang, Bing Qin, and Ting Liu. 2015.
\newblock Document modeling with gated recurrent neural network for sentiment
  classification.
\newblock In \emph{EMNLP}, pages 1422--1432.

\bibitem[{Wang and Manning(2012)}]{wang2012baselines}
Sida Wang and Christopher~D Manning. 2012.
\newblock Baselines and bigrams: Simple, good sentiment and topic
  classification.
\newblock In \emph{Proceedings of the 50th Annual Meeting of the Association
  for Computational Linguistics: Short Papers-Volume 2}, pages 90--94.
  Association for Computational Linguistics.

\bibitem[{Yang et~al.(2016)Yang, Yang, Dyer, He, Smola, and
  Hovy}]{yang2016hierarchical}
Zichao Yang, Diyi Yang, Chris Dyer, Xiaodong He, Alexander~J Smola, and
  Eduard~H Hovy. 2016.
\newblock Hierarchical attention networks for document classification.
\newblock In \emph{HLT-NAACL}, pages 1480--1489.

\bibitem[{Zhou et~al.(2015)Zhou, Sun, Liu, and Lau}]{zhou2015c}
Chunting Zhou, Chonglin Sun, Zhiyuan Liu, and Francis Lau. 2015.
\newblock A {C-LSTM} neural network for text classification.
\newblock \emph{arXiv preprint arXiv:1511.08630}.

\end{thebibliography}

\end{document}